\documentclass[11pt, letterpaper]{article}

\usepackage[margin=1in]{geometry}

\usepackage{epsfig,endnotes,xspace,url,diagbox}
\usepackage{epsfig,comment}
\usepackage{asymptote}

\usepackage{amsmath, amsthm, amsfonts}
\usepackage{newtxtext, newtxmath}
\usepackage{bbm}
\usepackage{enumerate}
\usepackage{varioref}
\usepackage{graphicx}
\usepackage{subfigure}
\usepackage{hyperref}
\usepackage{pifont}

\usepackage{caption}
\usepackage{subcaption}
\usepackage{float}
\usepackage{cleveref}
\usepackage{thmtools}
\usepackage{thm-restate}

\usepackage{epstopdf}
\usepackage{xcolor}
\usepackage{colortbl} 

\usepackage{multirow}
\usepackage{comment}
\usepackage{color}
\usepackage[utf8]{inputenc}
\usepackage{mathtools}

\usepackage{algorithm}
\usepackage{algorithmic}
\allowdisplaybreaks

\usepackage{wrapfig}
\usepackage{tabularx}
\usepackage{xspace}
\usepackage{cite}
\usepackage{booktabs}
\usepackage{tcolorbox}
\usepackage{enumitem}
\usepackage[strict]{changepage}
\usepackage{framed}
\usepackage{url}
\theoremstyle{plain}

\setlist[itemize]{leftmargin=*,noitemsep, topsep=0pt}

\setlist[enumerate]{leftmargin=*,noitemsep, topsep=0pt}

\newcommand{\eg}{\emph{e.g., }}

\newcommand{\etc}{\emph{etc.}}

\newcommand{\mypara}[1]{\smallskip\noindent\textbf{#1.} \xspace}

\newcommand{\mymethod}{\ensuremath{\mathsf{AutoVerifier}}\xspace}

\usepackage{todonotes}

\newcommand{\crossmark}{
  \ding{55}
}

\newtcolorbox{findingbox}[1][]{
  colback=gray!5,
  colframe=gray!60,
  fonttitle=\bfseries,
  title={#1},
  boxrule=0.5pt,
  arc=2pt,
  left=6pt, right=6pt, top=4pt, bottom=4pt
}

\definecolor{customshade}{rgb}{0.941, 0.937, 0.996}
\definecolor{darkblue}{rgb}{0.14,0.22,0.62}
\newenvironment{kkboxline}{%
  \MakeFramed{\advance\hsize-\width\FrameRestore}%
  \noindent\hspace{-4.55pt}%
  \begin{adjustwidth}{}{7pt}%
}{%
  \end{adjustwidth}\endMakeFramed%
}

\newtcolorbox{triplebox}[1][]{
  colback=blue!3,
  colframe=blue!40,
  fontupper=\small,
  fonttitle=\bfseries,
  title={#1},
  boxrule=0.5pt,
  arc=2pt,
  left=6pt, right=6pt, top=4pt, bottom=4pt
}

\title{\mymethod: An Agentic Automated Verification Framework Using Large Language Models}

\author{
    \\ 
    \begin{tabular}[t]{c}
        Yuntao Du\\
        \texttt{ytdu@purdue.edu}
    \end{tabular}
    \begin{tabular}[t]{c}
        Minh Dinh\\
        \texttt{dinh16@purdue.edu}
    \end{tabular}
    \begin{tabular}[t]{c}
        Kaiyuan Zhang\\
        \texttt{zhan4057@purdue.edu}
    \end{tabular}
    \begin{tabular}[t]{c}
        Ninghui Li\\
        \texttt{ninghui@purdue.edu}
    \end{tabular} \\ [2.5em]
    \textbf{TruSeLLM Team}\\[1em]
    \textbf{Purdue University}\\[1.5em]
    \textbf{Winner of 2025-2026 Radiance Technologies Innovation Bowl}\\[1em]
}

\date{}

\begin{document}

\begin{titlepage}
    \maketitle
    \thispagestyle{empty} 
\end{titlepage}

\begin{abstract}
Scientific and Technical Intelligence (S\&TI) analysis requires verifying complex technical claims across rapidly growing literature, where existing approaches fail to bridge the verification gap between surface-level accuracy and deeper methodological validity.
We present \mymethod, an LLM-based agentic framework that automates end-to-end verification of technical claims without requiring domain expertise.
\mymethod decomposes every technical assertion into structured claim triples of the form (Subject, Predicate, Object), constructing knowledge graphs that enable structured reasoning across six progressively enriching layers: corpus construction and ingestion, entity and claim extraction, intra-document verification, cross-source verification, external signal corroboration, and final hypothesis matrix generation.
We demonstrate \mymethod on a contested quantum computing claim, where the framework, operated by analysts with no quantum expertise, automatically identified overclaims and metric inconsistencies within the target paper, traced cross-source contradictions, uncovered undisclosed commercial conflicts of interest, and produced a final assessment.
These results show that structured LLM verification can reliably evaluate the validity and maturity of emerging technologies, turning raw technical documents into traceable, evidence-backed intelligence assessments.
\end{abstract}

\newpage
\pagenumbering{roman} 
\tableofcontents
\thispagestyle{empty} 

\newpage
\pagenumbering{arabic} 

\section{Introduction}
\label{sec:intro}

Scientific and Technical Intelligence (S\&TI) analysis evaluates technical developments to assess the capabilities, maturity, and trajectory of emerging technologies, guiding research investment, competitive positioning, and national security decisions~\cite{porter2005tech}.
In an era of rapid scientific publication, the key challenge is no longer information scarcity but \textit{signal validity}: distinguishing operational breakthroughs from incremental or purely theoretical contributions, verifying that reported results are methodologically sound, detecting overclaims where conclusions outpace supporting evidence, and identifying conflicts of interest or biased framing.
Existing approaches, such as Named Entity Recognition~\cite{li2022survey} and fact-checking systems~\cite{guo2022survey}, address individual facets of this problem but struggle with the complex, interdependent claims found in technical documents.
This leaves a \textit{verification gap}: the disconnect between surface-level factual accuracy and deeper methodological validity remains largely unaddressed.

Large Language Models (LLMs) offer a promising foundation to address this gap, but they are prone to hallucination and can produce plausible yet unsupported conclusions~\cite{ji2023survey, zhang2025siren}.
Harnessing their reasoning strengths while mitigating unconstrained inference requires a structured verification framework.

In this report, we propose \mymethod, an LLM-based agentic framework that bridges this verification gap.
Inspired by Palantir's Ontology~\cite{palantir2024ontology}, which models complex operational environments as objects, properties, and links, \mymethod imposes a similar discipline on S\&TI analysis.
Specifically, the framework first identifies \textit{entities}, the key actors and artifacts in technical literature, such as organizations, researchers, and algorithms.
Every technical assertion is then decomposed into \textit{claim triples} of the form \texttt{(Subject, Predicate, Object)}, where the subject and object are entities and the predicate captures their relationship.
These triples can be viewed as a knowledge graph~\cite{ji2021survey} that enables structured graph reasoning, rather than free-text inference, for multi-hop causal analysis~\cite{chen2020review}, contradiction detection, and discovery of implicit dependencies across sources.

\mymethod implements this approach through a six-layer pipeline that progressively enriches entities and claim triples, turning raw technical documents into verified, traceable intelligence assessments:

\begin{enumerate}
    \item \textbf{Corpus Construction \& Ingestion.} The pipeline collects technical artifacts from multiple source types, applies quality and bias-aware filtering, and ingests the selected documents into a searchable vector database that serves as the foundational evidence base.

    \item \textbf{Entity \& Claim Extraction.} From the ingested corpus, structured entities are identified and technical assertions are extracted as claim triples. Each claim is assigned a provenance level reflecting its evidentiary strength, along with normalized performance metrics.

    \item \textbf{Intra-Document Verification.} Each document is audited for internal consistency, enriching its claim triples with natural language inference verdicts, coherence flags, and overclaim annotations.

    \item \textbf{Cross-Source Verification.} Claims are triangulated across independent sources by retrieving related works, such as rebuttals, follow-up studies, and independent benchmarks, to detect contradictions and perform root cause analysis. All newly retrieved documents are processed through Layers~1 through Layer~3 to ensure they meet the same structural and evidentiary standards.

    \item \textbf{External Signal Corroboration.} Technical claims are further contextualized against non-academic signals, such as financial activity and partnerships, enriching each entity with a signal profile covering conflicts of interest, supply chain dependencies, and strategic positioning.

    \item \textbf{Hypothesis Matrix \& Reporting.} Finally, the pipeline aggregates all enriched triples and entity signals into a scored intelligence assessment, outputting a hypothesis matrix with technology maturity ratings and alpha signal detection.

\end{enumerate}

The framework is \textit{domain-agnostic}: while this report demonstrates its capabilities in quantum computing, the methodology applies to any emerging technology area where signal validity is critical.

\section{Methodology}
\label{sec:methodology}

\mymethod implements a six-layer verification pipeline in which each layer operates as an LLM-based agentic module, receiving structured outputs from preceding layers and passing enriched results to the next.

\subsection{Layer 1: Corpus Construction \& Ingestion}
\label{sec:layer1}

\mypara{Corpus Construction}
To build a reliable evidence base for downstream verification, \mymethod first assembles a comprehensive, bias-aware corpus of source material.
Given a query domain or target topic, the framework searches across diverse source types, such as academic research papers (\eg from arXiv and Google Scholar), patents, and author profiles.
Each collected source can then be scored for quality and reliability based on signals such as citation counts, affiliations, publication venues, and potential biases.

\mypara{Corpus Ingestion}
The collected documents undergo a three-stage processing pipeline to prepare raw, unstructured data for entity and claim extraction in subsequent layers:

\begin{enumerate}
    \item \textit{Text Extraction and Embedding.} Documents are converted from their native formats (PDF, HTML, \etc) into clean, searchable text with section boundaries preserved~\cite{lopez2009grobid}, then embedded and stored in a vector database for semantic retrieval~\cite{karpukhin2020dense}.

    \item \textit{Visual Asset Processing.} Figures, data plots, and architecture diagrams are extracted and paired with their captions and inline references. These visual assets are processed using vision-language models~\cite{liu2023visual} to generate semantic descriptions and extract data trends, which are also stored in the database.

    \item \textit{Metadata Alignment.} Structured metadata (\eg authors, affiliations, publication date) is extracted and stored for each document.
\end{enumerate}

\mypara{Deliverable}
It outputs a database with paper indexes, vector embeddings, and structured metadata, serving as the foundational evidence base.
This ingestion pipeline operates offline, and when new publications become available, only the new documents need to be ingested, without reconstructing the entire corpus.

\subsection{Layer 2: Entity \& Claim Extraction}
\label{sec:layer2}

This layer extracts the structured entities and claim triples that underpin downstream verification.

\mypara{Entity Extraction}
\mymethod uses structured prompts to instruct LLMs to extract key entities from the ingested documents, adapting to diverse technical terminology~\cite{wang2023gptner, xu2024llmie}.

\mypara{Claim Triple Construction}
Once entities are identified, \mymethod extracts the relationships and technical claims that connect them~\cite{wadhwa2023revisiting}, decomposing each assertion into the structured claim triples.

\mypara{Provenance Classification}
Not all claim triples carry the same evidential weight.
Each triple is annotated with a provenance level based on a five-tier hierarchy:

\begin{itemize}
    \item \textit{Level~1 (Experimental Data).} Claims grounded in physical system measurements or experiments.
    \item \textit{Level~2 (Simulation Result).} Claims supported by computational simulations.
    \item \textit{Level~3 (Theoretical Estimate).} Claims derived from analytical models or mathematical proofs.
    \item \textit{Level~4 (Citation of Another Work).} Claims referencing external results without independent verification.
    \item \textit{Level~5 (Author Assertion).} Claims stated without direct supporting evidence.
\end{itemize}

\mypara{Metric Normalization}
\mymethod extracts quantitative performance assertions from the claim triples and converts them into standardized units, identifying the measurement methodology behind each metric and flagging definitional discrepancies.

\mypara{Deliverable}
The output of this layer is a set of structured entities and provenance-classified claim triples, with normalized metrics for quantitative claims.
These outputs can be viewed as a knowledge graph, where entities serve as nodes and claim triples as edges, enabling graph-based reasoning in subsequent layers.

\subsection{Layer 3: Intra-Document Verification}
\label{sec:layer3}

With extracted entities and claim triples, this layer verifies individual documents for internal consistency.
It evaluates whether a document's own evidence supports its claim triples.

\mypara{Claim-Evidence Alignment}
For each claim triple, \mymethod locates supporting evidence within the source document and cross-references textual claims against visual assets using the stored multi-modal embeddings.
Each evidence passage is then classified as \textit{supports}, \textit{contradicts}, or \textit{neutral} through Natural Language Inference (NLI)-style reasoning~\cite{bowman2015large}.

\mypara{Methodology-Result Coherence}
\mymethod assesses whether a document's experimental methodology logically supports its reported results.
It compares the methodology and results sections across three dimensions: scope consistency, baseline fairness, and reproducibility.

\mypara{Overclaim Detection}
Overclaims are flagged where high-level statements exceed the supporting evidence~\cite{wright2021exaggeration}: conclusions that go beyond the data, claims that omit acknowledged limitations, and future projections presented as established results.

\mypara{Deliverable}
This layer enriches each claim with a source-text link, an NLI verdict, coherence flags, and overclaim annotations.
A document-level consistency score, currently defined as the proportion of claims supported by internal evidence, weights each source's contribution in the following cross-source analysis.

\subsection{Layer 4: Cross-Source Verification}
\label{sec:layer4}

This layer compares claim triples with independent sources, enriching them with cross-source agreement labels and consensus scores.

\mypara{Related Claim Discovery}
For each claim triple, \mymethod retrieves related works through three strategies: traversing citation networks to capture foundational studies, rebuttals, and follow-up work, performing semantic similarity searches over stored embeddings~\cite{karpukhin2020dense} to find topically related works without direct citation links, and conducting entity-based graph traversal to find documents sharing core entities (\eg organizations and algorithms).
Beyond direct rebuttals and follow-ups, semantic similarity search can also retrieve evaluation frameworks and surveys that provide external rubrics for assessing claims.
All newly discovered sources are processed through Layers~1, 2, and 3, creating a unified, verified evidence base for cross-source comparison.

\mypara{Claim Alignment and Contradiction Analysis}
Because different authors may phrase the same claim differently, \mymethod uses NLI-style reasoning~\cite{bowman2015large} to classify each pair of claim triples as matched, partially overlapping, or unrelated.
For matched claims, \mymethod performs two further checks:
\begin{itemize}
    \item \textit{Citation Fidelity.} When Source~A cites Source~B to support a claim, \mymethod retrieves Source~B's original claims and verifies whether they actually support Source~A's assertion. This detects ``citation distortion'', where cited findings are inadvertently or strategically exaggerated~\cite{greenberg2009citation}.
    \item \textit{Contradiction Root Cause Analysis.} When claims from different sources are conflicted, \mymethod uses Chain-of-Thought prompting~\cite{wei2022chain} to trace through the methodological details of each source, identifying why the results disagree. Discrepancies are classified as arising from methodological differences, incompatible experimental conditions, differing benchmark datasets, and others.
\end{itemize}

\mypara{Source Independence Assessment}
Source independence is evaluated using bibliometric analysis~\cite{zupic2015bibliometric}, scores of author overlap, shared institutional affiliations, and citation graph proximity.
Sources with high overlap are down-weighted, and the document-level consistency score from Layer~3 is incorporated so that internally inconsistent sources contribute less.
Each source receives a High, Medium, or Low independence rating that weights its contribution to the cross-source consensus score.

\mypara{Deliverable}
This layer enriches each claim triple with cross-source agreement labels (Corroborates, Contradicts, or Misrepresents, with contradictions annotated by root cause) and a consensus score weighted by source independence and internal consistency.

\subsection{Layer 5: External Signal Corroboration}
\label{sec:layer5}

With claim triples now verified both internally and across sources, this layer contextualizes them against non-academic signals not captured in the technical literature.

\mypara{Financial Profile and Conflict of Interest Detection}
\mymethod constructs a financial profile for each entity using public sources such as SEC filings and funding databases.
Spending patterns are classified as Capital Expenditure (CapEx) or Operating Expenditure (OpEx) to assess whether an entity is making long-term hardware investments or relying on third-party infrastructure.
Conflicts of interest are detected by linking author names to corporate officer records, flagging cases where an author holds a financial or leadership stake in a commercially tied entity.

\mypara{Supply Chain Dependency Mapping}
Technical viability frequently hinges on external dependencies that are rarely disclosed in academic papers. 
\mymethod performs multi-hop reasoning~\cite{chen2020review} over the extracted entities and claim triples, chaining together hardware prerequisites and manufacturing relationships to uncover complex supply chain dependencies.

\mypara{Strategic Signal Integration}
News and press releases are parsed~\cite{xu2024llmie} to build a chronological timeline of strategic events for each entity, including partnerships, acquisitions, and funding rounds. 
By analyzing temporal correlations between stakeholders, such as hardware manufacturers and research institutions, the system distinguishes sustained R\&D investment from superficial, announcement-driven posturing.

\mypara{Deliverable}
This layer enriches each entity with a signal profile (financial summary, conflict-of-interest flags, supply chain dependencies, and strategic positioning), which will be used for the final assessment.

\subsection{Layer 6: Hypothesis Matrix \& Reporting}
\label{sec:layer6}

This final layer produces the intelligence assessment.
It aggregates each claim triple's provenance level (Layer~2), consistency verdict (Layer~3), cross-source consensus (Layer~4), and entity-level signals (Layer~5) into a unified evidence profile.
From this profile, it generates testable hypotheses via Chain-of-Thought prompting~\cite{wei2022chain} and adversarial counter-hypotheses by exploring alternative reasoning paths~\cite{yao2023tree}.
Confidence is estimated as a qualitative level (Low, Medium, High) based on semantic entropy~\cite{kuhn2023semantic} across multiple generations and agreement among independent LLMs.
The output is a \textit{hypothesis matrix} where each row contains a hypothesis, supporting evidence, cross-source consistency, confidence level, counter-hypotheses, and a final label of \textit{Supported}, \textit{Needs Review}, or \textit{Likely Hallucination}, accompanied by a \textit{technology maturity assessment} that flags \textit{alpha signals} where all layers converge positively.

\section{Execution}
\label{sec:execution}

\mypara{Case Study and Target Query}
We apply \mymethod to quantum computing, where our team has no prior expertise, providing a rigorous test of the framework's ability to operate without domain-specific guidance.
The target paper~\cite{chandarana2025runtime} proposes Bias-Field Digitized Counterdiabatic Quantum Optimization (BF-DCQO), a quantum algorithm developed by researchers affiliated with Kipu Quantum GmbH.
The paper claims that BF-DCQO achieves a runtime advantage on Higher-Order Unconstrained Binary Optimization (HUBO) problems over classical solvers, specifically Simulated Annealing (SA) and IBM's CPLEX exact optimizer, when executed on IBM's 156-qubit Heron Quantum Processing Unit (QPU).
This is exactly the type of contested, high-impact claim that \mymethod is designed to evaluate: a reported operational breakthrough in an emerging technology domain.
We formulate the following target query:

\begin{center}
\textit{Does the paper ``Runtime Quantum Advantage with Digital Quantum Optimization'' achieve true runtime quantum advantage?}
\end{center}

Starting from this paper~\cite{chandarana2025runtime}, \mymethod automatically expands its analysis through the full six-layer pipeline, requiring no further manual intervention.

\mypara{Implementation}
Since each layer in \mymethod relies on declarative, domain-agnostic prompts and structured deliverables, it supports different orchestration strategies.
We implemented and evaluated two approaches.
The first is a \textit{multi-LLM sequential pipeline}, a Python application that uses LiteLLM~\cite{litellm2024} to route each layer to a different model.
Layers~1 and Layer~2 use Perplexity Sonar~\cite{perplexity2025sonar} for web search and retrieval, while Layers~3 through Layer~6 use Google Gemini Pro~\cite{google2025gemini} for multi-modal reasoning and large-context processing.
The second is a \textit{single-agent framework} using agentic AI tools such as Claude Code~\cite{anthropic2025claudecode} or OpenClaw~\cite{steinberger2025openclaw}.
A single LLM agent receives the target query and all six layer specifications in one prompt, then works through the pipeline by calling built-in tools: web search, document retrieval, and execution.
The results in this report were produced using the Claude Code implementation.

\section{Results}
\label{sec:results}

We present representative outputs from each pipeline layer for the case study described above.

\subsection{Corpus Construction and Claim Extraction}

\begin{kkboxline}
\begin{itemize}[leftmargin=*, nosep]
    \item \textbf{Corpus.} 11 sources collected across 5 research groups: 1 target paper, 4 Kipu-authored supporting papers (all sharing $\geq$4/6 authors with the target paper), 3 independent rebuttals, 2 independent benchmarks, and 1 external evaluation framework.
    From the target paper, 17 entities and 20 provenance-classified claim triples were extracted (see Table~\ref{tab:extraction} for examples).
    \item \textbf{Execution classification.} \mymethod determined that the target paper ran on a physical quantum processor with high confidence, based on named hardware backends (\texttt{ibm\_marrakesh}, \texttt{ibm\_kingston}) and circuit elements matched to heavy-hex connectivity.
    \item \textbf{Runtime definition.} Both quantum and classical runtimes exclude significant overhead, making direct comparison unreliable (Table~\ref{tab:metrics}).
    However, the impact is asymmetric: the excluded transpilation overhead on the quantum side (${\sim}2\,$s) is comparable to BF-DCQO's total reported runtime of 0.2--2.2\,s, so including it would roughly double the quantum wall-clock time and largely eliminate the claimed speedup.
    The excluded initialization overhead on the classical side (${\sim}1.65\,$s) is a smaller fraction of typical SA runtimes and has less effect on the comparison.
\end{itemize}
\end{kkboxline}

\begin{table}[h]
\centering
\small
\caption{Examples of extracted claim triples with provenance classification and intra-document verification.}
\label{tab:extraction}
\begin{tabularx}{\textwidth}{llXlcl}
\toprule
\textbf{Subject} & \textbf{Predicate} & \textbf{Object} & \textbf{Provenance} & \textbf{Verdict} & \textbf{Source} \\
\midrule
BF-DCQO & executed-on     & IBM Heron (Marrakesh/Kingston)            & Lv.~1 Exp. & Supports & Sec.~IV, p.5 \\
BF-DCQO & outperforms (time-to-result) & SA ($>$3.5$\times$) / CPLEX (up to 80$\times$) & Lv.~1 Exp. & Partial & Sec.~V \\
BF-DCQO & projected-to-achieve & ``several orders of magnitude''    & Lv.~5 Assert.    & Overclaim & Abstract, p.1 \\
BF-DCQO & requires        & classical SA pre- and post-processing   & Lv.~1 Exp. & Supports & Sec.~III.C, Fig.~4 \\
\bottomrule
\end{tabularx}
\end{table}

\begin{table}[h]
\centering
\small
\caption{Normalized metric comparison: quantum vs.\ classical runtime definitions.}
\label{tab:metrics}
\begin{tabularx}{\textwidth}{lXXc}
\toprule
\textbf{Metric} & \textbf{Quantum (BF-DCQO)} & \textbf{Classical (SA/CPLEX)} & \textbf{Comparable?} \\
\midrule
Runtime definition & $T_{\text{CPU}} + T_{\text{QPU}}$ (partial) & $T_{\text{sweep}}$ formula (partial) & \crossmark \\
Included overhead  & CPU pre-/post-processing, QPU gates & Sweep computation only & \crossmark \\
Excluded overhead  & Transpilation (${\sim}2\,$s), readout, queuing & Initialization (${\sim}1.65\,$s), solver setup & \crossmark \\
\bottomrule
\end{tabularx}
\end{table}

\subsection{Intra-Document Verification}
 
\begin{kkboxline}
\begin{itemize}[leftmargin=*, nosep]
    \item \textbf{Consistency score.} Of the 20 claim triples, 6 are fully supported by internal evidence (30\%), 8 are partially supported, 3 are overclaims, and 3 are neutral or descriptive
     (Table~\ref{tab:overclaims} shows some examples).
    \item \textbf{Overclaim escalation.} The methods sections (Sec.~III--IV) use appropriate hedging (``best-performing instance''), but the abstract and conclusion drop these qualifiers, presenting instance-specific results as general findings (``achieving runtime quantum advantage,'' ``industrial-scale optimization'').
    \item \textbf{Baseline and metric concerns.} CPLEX is benchmarked on a single CPU thread, with the reported ``$80\times$'' enhancement being a single outlier (median ${\sim}5$--$7\times$).
    The ``several orders of magnitude'' projection has no supporting analysis, and no industrial problem is tested despite claims of industrial-scale applicability (Table~\ref{tab:overclaims}).
    \item \textbf{Hybrid framing.} BF-DCQO requires classical SA for both pre-processing (warm start) and post-processing (zero-temperature refinement), making it a hybrid classical-quantum-classical workflow (Sec.~III.C, Fig.~4).
    The hybrid design is a valid engineering choice, but labeling the end-to-end speedup ``quantum advantage'' attributes the performance gain to the QPU without isolating its contribution from the classical steps.
\end{itemize}
\end{kkboxline}
 
\begin{table}[h]
\centering
\small
\caption{Overclaim detection: claims vs.\ supporting evidence in the target paper.}
\label{tab:overclaims}
\begin{tabularx}{\textwidth}{llXX}
\toprule
\textbf{Issue} & \textbf{Severity} & \textbf{Claim} & \textbf{Evidence} \\
\midrule
Overgeneralization & Moderate & ``achieving runtime quantum advantage'' (Abstract) & Demonstrated only on ``hardest instance out of a pool of 250'' (Sec.~IV.C) \\
Extreme-value reporting & Moderate & ``up to a factor of 80'' speedup (Sec.~V) & Single maximum outlier; median enhancement ${\sim}5$--$7\times$ across instances \\
Projection as result & Severe & ``several orders of magnitude'' (Abstract) & Prefaced with ``we expect''; no supporting data or analysis in body \\
Scope inflation & Severe & ``industrial-scale optimization problems'' (Sec.~V) & Only synthetic HUBO tested; $N{\leq}156$; no industrial problem formulated \\
\bottomrule
\end{tabularx}
\end{table}

\subsection{Cross-Source Verification}

\begin{kkboxline}
\begin{itemize}[leftmargin=*, nosep]
    \item \textbf{Consensus.} QPU execution is confirmed by all sources, but runtime advantage is contradicted by every independent evaluation (Table~\ref{tab:contradiction} shows some examples).
    \item \textbf{Three root causes.}
    The target paper~\cite{chandarana2025runtime} and its critics disagree on three methodological dimensions:
    (1)~\textit{Runtime definition}: end-to-end wall-clock timing~\cite{tuziemski2025runtime} eliminates the advantage.
    (2)~\textit{Baseline selection}: stronger solvers (the Simulated Bifurcation Machine~\cite{tuziemski2025runtime}, enhanced parallel tempering, and GPU-accelerated solvers~\cite{chandarana2026quest}) eliminate it entirely.
    (3)~\textit{Statistical sampling}: averaging over instances rather than cherry-picking reverses the conclusion~\cite{tuziemski2025runtime}.
    \item \textbf{BF-Null falsification.} A D-Wave rebuttal~\cite{farre2025comparing} replaced the QPU with a trivial classical sweep (``BF-Null'') and obtained comparable solution quality, providing direct evidence that the classical components of the pipeline drive performance.
    \item \textbf{Self-correction signal.} Kipu's own publication trajectory retreats from the original claim: the target paper (May 2025) claims ``runtime quantum advantage,'' the follow-up~\cite{chandarana2025hsqc} (October 2025) reframes as ``hybrid sequential quantum computing,'' and the latest benchmark~\cite{chandarana2026quest} (March 2026) acknowledges classical solvers ``reach or surpass'' the hybrid workflow.
    \item \textbf{Zero independent corroboration.} All 4 supporting papers are Kipu-led. The 2 independent evaluations that directly assess BF-DCQO~\cite{tuziemski2025runtime,farre2025comparing} both contradict the advantage claim.
    \item \textbf{Keystone evaluation.} Huang et al.~\cite{huang2025vast} propose five keystone properties for credible quantum advantage: Predictability, Typicality, Robustness, Verifiability, and Usefulness. BF-DCQO meets 0/5 fully: Typicality fails (cherry-picked instances), Robustness fails (advantage vanishes against stronger baselines and BF-Null), and Verifiability is limited (proprietary code, no data release).
\end{itemize}
\end{kkboxline}

\begin{table}[h]
\centering
\small
\caption{Cross-source contradiction: target paper vs.\ two independent evaluations.}
\label{tab:contradiction}
\begin{tabularx}{\textwidth}{lXXX}
\toprule
\textbf{Dimension} & \textbf{Target Paper~\cite{chandarana2025runtime}} & \textbf{Rebuttal~\cite{tuziemski2025runtime}} & \textbf{BF-Null Control~\cite{farre2025comparing}} \\
\midrule
Advantage claimed?  & Yes                          & No & No (quantum component minimal) \\
QPU execution?      & Confirmed                    & Confirmed (not disputed) & N/A (tests contribution, not execution) \\
Runtime definition  & $T_{\text{CPU}} + T_{\text{QPU}}$ (partial) & End-to-end wall-clock & Full pipeline wall-clock \\
Classical baselines & SA (fixed sweeps), CPLEX (1 thread) & SBM, optimized SA & SA pre/post-processing only (QPU removed) \\
Independence        & Low (commercially interested) & High (no affiliation) & Medium (competitor, but reproducible control) \\
\bottomrule
\end{tabularx}
\end{table}

\subsection{External Signal Corroboration}

\begin{kkboxline}
\begin{itemize}[leftmargin=*, nosep]
    \item \textbf{Verified COI.} All six authors are employed by Kipu Quantum.
    Co-founder Enrique Solano (CEO and Co-Founder~\cite{kipuquantum2025team}) holds an equity stake.
    BF-DCQO is their commercial product (``Iskay Quantum Optimizer''), listed in IBM's Qiskit Functions Catalog~\cite{ibm2025iskay}.
    No competing interests are disclosed.
    \item \textbf{Financial incentive.} Kipu raised ${\sim}$\texteuro13.5M in disclosed seed funding~\cite{kipuquantum2022seed,kipuquantum2023seed} and acquired Anaqor AG's quantum computing platform~\cite{kipuquantum2024anaqor}. The company is OpEx-dominant with no hardware manufacturing, making BF-DCQO performance claims its primary value proposition.
    \item \textbf{Publication-product timing.} The Iskay product launched on IBM's marketplace in March 2025~\cite{ibm2025iskay}.
    The ``runtime quantum advantage'' paper followed two months later (May 2025).
    \item \textbf{IBM conflict web.} IBM simultaneously provides the QPU hardware, owns the classical baseline (CPLEX), hosts the Iskay product, and co-authors the latest benchmark~\cite{chandarana2026quest} via an IBM Research scientist who also validated T1's CPLEX benchmark~\cite{chandarana2025runtime}.
    \item \textbf{Proprietary black box.} Kipu has released no source code, no instance data, and no reproducible implementation: ``The exact implementation of the algorithm \ldots\ is proprietary to Kipu Quantum''~\cite{chandarana2026quest}.
    \item \textbf{Temporal signal.} Kipu published a follow-up~\cite{chandarana2025hsqc} using softer language (``runtime quantum-advantage level'') within days of the independent rebuttal~\cite{tuziemski2025runtime} (both October 2025).
\end{itemize}
\end{kkboxline}

The verified COI and hardware dependency findings above were identified through multi-hop reasoning over extracted entities:

\begin{triplebox}[Multi-Hop Reasoning Chains for External Signal Corroboration]
\textit{COI Discovery:} \\
\texttt{Solano} (co-author)
  $\xrightarrow{\textit{\scriptsize co-founded}}$ \texttt{Kipu Quantum}
  $\xrightarrow{\textit{\scriptsize sells}}$ \texttt{Iskay Optimizer}
  $\xrightarrow{\textit{\scriptsize implements}}$ \texttt{BF-DCQO} (evaluated algorithm) \\[6pt]
\textit{IBM Conflict Web:} \\
\texttt{IBM}
  $\xrightarrow{\textit{\scriptsize provides}}$ \texttt{Heron QPU} (hardware),\quad
  $\xrightarrow{\textit{\scriptsize owns}}$ \texttt{CPLEX} (classical baseline),\quad
  $\xrightarrow{\textit{\scriptsize hosts}}$ \texttt{Iskay} (product),\quad
  $\xrightarrow{\textit{\scriptsize co-authors}}$ \texttt{S4} (benchmark) \\[6pt]
\textit{Supply Chain:} \\
\texttt{BF-DCQO}
  $\xrightarrow{\textit{\scriptsize requires}}$ \texttt{IBM Heron}
  $\xrightarrow{\textit{\scriptsize topology}}$ \texttt{Heavy-hex lattice}
  $\xrightarrow{\textit{\scriptsize manufactured by}}$ \texttt{IBM}
\end{triplebox}

\subsection{Hypothesis Matrix and Final Assessment}

\begin{kkboxline}
\begin{itemize}[leftmargin=*, nosep]
    \item \textbf{Split verdict.} \mymethod assigns high confidence to the finding that the technology works on real hardware (Technology Readiness Level (TRL)~\cite{mankins1995trl} 4--5), but rates the claimed performance advantage as \textit{Likely Hallucination} because it does not survive any independent evaluation~\cite{tuziemski2025runtime,farre2025comparing} and is implicitly retracted by the claimant's own subsequent work~\cite{chandarana2026quest}.
    \item \textbf{Three root causes.} The most probable explanations for the claimed speedup are a measurement artifact (transpilation overhead excluded), weak classical baselines (advantage disappears against SBM, PT+, ABS3), and a possible quantum-null result (BF-Null classical replacement matches performance~\cite{farre2025comparing}).
    \item \textbf{Confidence validation.} Supported findings (\eg QPU execution) show low semantic entropy (0.12) and full multi-model agreement (3/3), while disputed findings (\eg runtime advantage) show high entropy (0.68) and split agreement (1/3).
    \item \textbf{Alpha signal detection.} No dimension shows convergence across all pipeline layers toward a genuine operational breakthrough.
    The closest positive signal, that BF-DCQO produces fast approximate solutions, does not constitute an advantage since classical solvers achieve comparable results.
    \item \textbf{Overall maturity.} TRL~4--5 (Experimental to Early Commercial). The algorithm is functional and commercially deployed (Iskay), but its core value proposition is not independently established.
\end{itemize}
\end{kkboxline}


\begin{table}[h]
\centering
\footnotesize
\caption{Example hypothesis matrix for the runtime quantum advantage claim.}
\label{tab:hypothesis}
\begin{tabularx}{\textwidth}{Xllcl>{\centering\arraybackslash}p{1.1cm}}
\toprule
\textbf{Hypothesis} & \textbf{Evidence} & \textbf{Cross-Source} & \textbf{Confidence} & \textbf{Alternatives} & \textbf{Status} \\
\midrule
Runtime advantage is genuine & \cite{chandarana2025runtime,chandarana2025hsqc} & Contradicted~\cite{tuziemski2025runtime,farre2025comparing} & High entropy & Measurement artifact & Likely Hallucination \\
QPU contributes to performance & \cite{chandarana2025runtime} & Contradicted~\cite{farre2025comparing} & High entropy & Classical iteration suffices & Needs Review \\
Advantage is a measurement artifact & \cite{tuziemski2025runtime,chandarana2026quest} & Supported & Low entropy & Hybrid accounting & Supported \\
Real but incremental progress & \cite{chandarana2025runtime,tuziemski2025runtime,chandarana2026quest} & Consensus & Low entropy & Next-gen up-scaling & Supported \\
\bottomrule
\end{tabularx}
\end{table}

\section{Discussion}
\label{sec:discussion}

The quantum computing case study shows that structured LLM verification is essential for S\&TI analysis.
Using \mymethod’s six-layer pipeline, we validated complex technical claims in a domain where the team had no expertise, demonstrating its effectiveness in addressing the verification gap.

\mypara{Moving Beyond Summarization to Intra-Document Verification}
A naive LLM tool would likely ingest the target paper and confidently report that ``runtime quantum advantage has been achieved,'' simply trusting the claims in the abstract.
\mymethod avoids this hallucination trap by imposing structural discipline: intra-document verification (Layer~3) audits each document's claims against its own evidence using extracted claim triples, catching internal inconsistencies that unconstrained LLM inference would miss.
For example, the system compared the abstract's claim of ``several orders of magnitude'' advantage against the body text and correctly identified it as a projection unsupported by experimental data (Table~\ref{tab:overclaims}).
It further detected an overclaim escalation pattern: the methods sections use appropriate hedging (``best-performing instance,'' ``can potentially''), while the abstract and conclusion drop these qualifiers and extend to unsupported claims of ``industrial-scale optimization.''
This pattern (cautious methods, assertive framing) is a structural signal of strategic overclaiming.
\mymethod does not take author assertions at face value, but rather audits them against their own underlying evidence.

\mypara{Root-Cause Analysis of Cross-Source Contradictions}
When multiple sources conflict, \mymethod goes beyond flagging disagreement to perform root-cause analysis (Layer~4), assessing who is making each claim, whether the sources are truly independent, and why they disagree.
In the case study, the pipeline traced the contradiction to three distinct root causes:
(1)~\textit{incompatible runtime definitions} (partial QPU time vs.\ end-to-end wall-clock),
(2)~\textit{baseline dependency}, where the advantage vanishes against every stronger solver introduced by any source (SBM~\cite{tuziemski2025runtime}, PT+, ABS3~\cite{chandarana2026quest}), and
(3)~a \textit{possible null result for the quantum component}, where a D-Wave rebuttal~\cite{farre2025comparing} (found through citation-chain retrieval) replaced the QPU with a trivial classical sweep and obtained comparable quality, raising a question the target paper never asks: \textit{does the quantum component actually contribute?}
Identifying these required multiple layers working in concert: Layer~2 metric normalization flagged the incompatible definitions, Layer~3 detected the single-thread CPLEX configuration, and Layer~4 cross-source retrieval discovered the BF-Null control experiment that the original authors did not consider.

\mypara{Contextualizing Science with Commercial Reality}
\mymethod enriches technical analysis with external signals (Layer~5), contextualizing claims within commercial reality using financial data, patent records, and corporate documentation.
In the case study, \mymethod discovered that the target paper's co-author co-founded the commercial entity selling the evaluated algorithm~\cite{kipuquantum2025team,ibm2025iskay}.
This finding reframes the ``runtime advantage'' claim from a purely academic finding into a strategic commercial narrative.
The publication-product timeline reinforces this: the Iskay product launched on IBM's marketplace in March 2025~\cite{ibm2025iskay}, and the paper claiming ``runtime quantum advantage'' followed two months later, consistent with scientific claims timed to support commercial positioning.
Multi-hop reasoning over the entities and triples also uncovered IBM's quadruple conflict: IBM simultaneously provides the QPU hardware, owns the classical baseline (CPLEX), hosts the commercial product (Iskay), and co-authors the benchmark~\cite{chandarana2026quest}, meaning no link in this chain provides independent validation.
Finally, the self-correction signal provided the most compelling evidence: Kipu's own publication trajectory retreated from ``runtime quantum advantage'' (May 2025) to ``hybrid sequential quantum computing''~\cite{chandarana2025hsqc} (October 2025) to acknowledging that classical solvers ``reach or surpass'' the hybrid workflow~\cite{chandarana2026quest} (March 2026).
This evidence came from the claimant, making it more compelling than external critiques alone.
These findings show that alternative sources are not supplementary to technical analysis but provide independently verifiable evidence that academic sources lack.

\section{Conclusion and Recommendation}
\label{sec:conclusion}

In this report, we show that \mymethod is effective for S\&TI analysis.
By enforcing a rigorous, six-layer verification pipeline, the framework bridges the gap between naive summarization and deep methodological verification.
The system effectively dismantled a contested ``runtime advantage'' claim by isolating metric discrepancies, reversing a misleading literature consensus through source independence weighting, confirming undisclosed commercial dependencies, and mapping the full hardware supply chain.
\mymethod proves that LLMs can be structured to reliably evaluate the validity and maturity of emerging technologies without requiring prior domain expertise from the analyst.

Two directions would strengthen \mymethod in future iterations.
First, packaging each layer as a reusable \textit{LLM agent skill}~\cite{xu2026agentskills, anthropic2025claudecodeskills} would let analysts customize the pipeline for new domains by selecting, reordering, or extending layers with domain-specific checks.
Second, the system currently produces static, point-in-time assessments.
Shifting to continuous monitoring that automatically updates and adds entity profiles and claim triples as new publications and financial disclosures appear would maintain this information as a living intelligence resource.

\newpage
\bibliographystyle{plain} 
\bibliography{ref} 

@book{porter2005tech,
  author    = {Porter, Alan L. and Cunningham, Scott W.},
  title     = {Tech Mining: Exploiting New Technologies for Competitive Advantage},
  publisher = {John Wiley \& Sons},
  year      = {2005},
  isbn      = {978-0-471-47567-5}
}

@article{greenberg2009citation,
  author  = {Greenberg, Steven A.},
  title   = {How Citation Distortions Create Unfounded Authority: Analysis of a Citation Network},
  journal = {BMJ},
  year    = {2009},
  volume  = {339},
  pages   = {b2680},
  doi     = {10.1136/bmj.b2680}
}

@article{huang2025vast,
  author  = {Huang, Hsin-Yuan and Choi, Soonwon and McClean, Jarrod R. and Preskill, John},
  title   = {The Vast World of Quantum Advantage},
  journal = {arXiv preprint arXiv:2508.05720},
  year    = {2025}
}

@misc{palantir2024ontology,
  author       = {{Palantir Technologies}},
  title        = {Ontology Overview},
  year         = {2024},
  howpublished = {\url{https://www.palantir.com/docs/foundry/ontology/overview}},
  note         = {Accessed: 2025-03-23},
}

@article{zhang2025siren,
  title={Siren's Song in the AI Ocean: A Survey on Hallucination in Large Language Models},
  author={Zhang, Yue and Li, Yafu and Cui, Leyang and Cai, Deng and Liu, Lemao and Fu, Tingchen and Huang, Xinting and Zhao, Enbo and Zhang, Yu and Chen, Yulong and others},
  journal={Computational Linguistics},
  pages={1--46},
  year={2025},
}

@article{wei2022chain,
  title={Chain-of-thought prompting elicits reasoning in large language models},
  author={Wei, Jason and Wang, Xuezhi and Schuurmans, Dale and Bosma, Maarten and Xia, Fei and Chi, Ed and Le, Quoc V and Zhou, Denny and others},
  journal={Advances in neural information processing systems},
  volume={35},
  pages={24824--24837},
  year={2022}
}

@article{ji2023survey,
  title={Survey of hallucination in natural language generation},
  author={Ji, Ziwei and Lee, Nayeon and Frieske, Rita and Yu, Tiezheng and Su, Dan and Xu, Yan and Ishii, Etsuko and Bang, Ye Jin and Madotto, Andrea and Fung, Pascale},
  journal={ACM computing surveys},
  volume={55},
  number={12},
  pages={1--38},
  year={2023},
  publisher={ACM New York, NY}
}

@article{ji2021survey,
  title={A survey on knowledge graphs: Representation, acquisition, and applications},
  author={Ji, Shaoxiong and Pan, Shirui and Cambria, Erik and Marttinen, Pekka and Yu, Philip S},
  journal={IEEE transactions on neural networks and learning systems},
  volume={33},
  number={2},
  pages={494--514},
  year={2021},
  publisher={IEEE}
}

@article{chen2020review,
  title={A review: Knowledge reasoning over knowledge graph},
  author={Chen, Xiaojun and Jia, Shengbin and Xiang, Yang},
  journal={Expert systems with applications},
  volume={141},
  pages={112948},
  year={2020},
  publisher={Elsevier}
}

@article{guo2022survey,
  title={A Survey on Automated Fact-Checking},
  author={Guo, Zhijiang and Schlichtkrull, Michael and Vlachos, Andreas},
  journal={Transactions of the Association for Computational Linguistics},
  volume={10},
  pages={178--206},
  year={2022},
  publisher={MIT Press}
}

@article{li2022survey,
  title={A Survey on Deep Learning for Named Entity Recognition},
  author={Li, Jing and Sun, Aixin and Han, Jianglei and Li, Chenliang},
  journal={IEEE Transactions on Knowledge and Data Engineering},
  volume={34},
  number={1},
  pages={50--70},
  year={2022},
  doi={10.1109/TKDE.2020.2981314}
}

@inproceedings{yao2023tree,
  title={Tree of Thoughts: Deliberate Problem Solving with Large Language Models},
  author={Yao, Shunyu and Yu, Dian and Zhao, Jeffrey and Shafran, Izhak and Griffiths, Thomas L. and Cao, Yuan and Narasimhan, Karthik},
  booktitle={Advances in Neural Information Processing Systems},
  volume={36},
  year={2023}
}

@inproceedings{kuhn2023semantic,
  title={Semantic Uncertainty: Linguistic Invariances for Uncertainty Estimation of Large Language Models},
  author={Kuhn, Lorenz and Gal, Yarin and Farquhar, Sebastian},
  booktitle={International Conference on Learning Representations},
  year={2023}
}

@inproceedings{karpukhin2020dense,
  title={Dense Passage Retrieval for Open-Domain Question Answering},
  author={Karpukhin, Vladimir and Oguz, Barlas and Min, Sewon and Lewis, Patrick and Wu, Ledell and Edunov, Sergey and Chen, Danqi and Yih, Wen-tau},
  booktitle={Proceedings of the 2020 Conference on Empirical Methods in Natural Language Processing (EMNLP)},
  pages={6769--6781},
  year={2020},
  publisher={Association for Computational Linguistics},
  doi={10.18653/v1/2020.emnlp-main.550}
}

@inproceedings{liu2023visual,
  title={Visual Instruction Tuning},
  author={Liu, Haotian and Li, Chunyuan and Wu, Qingyang and Lee, Yong Jae},
  booktitle={Advances in Neural Information Processing Systems},
  volume={36},
  year={2023}
}

@inproceedings{lopez2009grobid,
  title={{GROBID}: Combining Automatic Bibliographic Data Recognition and Term Extraction for Scholarship Publications},
  author={Lopez, Patrice},
  booktitle={Research and Advanced Technology for Digital Libraries, 13th European Conference, ECDL 2009},
  pages={473--474},
  year={2009},
  publisher={Springer},
  series={Lecture Notes in Computer Science},
  volume={5714},
  doi={10.1007/978-3-642-04346-8\_62}
}

@inproceedings{wang2023gptner,
  title={{GPT-NER}: Named Entity Recognition via Large Language Models},
  author={Wang, Shuhe and Sun, Xiaofei and Li, Xiaoya and Ouyang, Rongbin and Wu, Fei and Zhang, Tianwei and Li, Jiwei and Wang, Guoyin},
  booktitle={Findings of the Association for Computational Linguistics: NAACL 2025},
  pages={4257--4275},
  year={2023},
  note={arXiv preprint arXiv:2304.10428}
}

@article{xu2024llmie,
  title={Large Language Models for Generative Information Extraction: A Survey},
  author={Xu, Derong and Chen, Wei and Peng, Wenjun and Zhang, Chao and Xu, Tong and Zhao, Xiangyu and Wu, Xian and Zheng, Yefeng and Wang, Yang and Chen, Enhong},
  journal={Frontiers of Computer Science},
  volume={18},
  number={6},
  pages={186357},
  year={2024},
  publisher={Springer}
}

@inproceedings{wadhwa2023revisiting,
  title={Revisiting Relation Extraction in the era of Large Language Models},
  author={Wadhwa, Somin and Amir, Silvio and Wallace, Byron},
  booktitle={Proceedings of the 61st Annual Meeting of the Association for Computational Linguistics (Volume 1: Long Papers)},
  pages={15566--15589},
  year={2023},
  address={Toronto, Canada},
  publisher={Association for Computational Linguistics},
  doi={10.18653/v1/2023.acl-long.868}
}

@inproceedings{bowman2015large,
  title={A Large Annotated Corpus for Learning Natural Language Inference},
  author={Bowman, Samuel R. and Angeli, Gabor and Potts, Christopher and Manning, Christopher D.},
  booktitle={Proceedings of the 2015 Conference on Empirical Methods in Natural Language Processing},
  pages={632--642},
  year={2015},
  publisher={Association for Computational Linguistics},
  doi={10.18653/v1/D15-1075}
}

@inproceedings{wright2021exaggeration,
  title={Semi-Supervised Exaggeration Detection of Health Science Press Releases},
  author={Wright, Dustin and Augenstein, Isabelle},
  booktitle={Proceedings of the 2021 Conference on Empirical Methods in Natural Language Processing},
  pages={10824--10836},
  year={2021},
  publisher={Association for Computational Linguistics},
  doi={10.18653/v1/2021.emnlp-main.845}
}

@article{zupic2015bibliometric,
  author    = {Zupic, Ivan and {\v{C}}ater, Toma{\v{z}}},
  title     = {Bibliometric Methods in Management and Organization},
  journal   = {Organizational Research Methods},
  volume    = {18},
  number    = {3},
  pages     = {429--472},
  year      = {2015},
  doi       = {10.1177/1094428114562629},
  publisher = {SAGE Publications}
}

@article{chandarana2025runtime,
  title={Runtime Quantum Advantage with Digital Quantum Optimization},
  author={Chandarana, Pranav and Cadavid, Alejandro Gomez and Romero, Sebasti{\'a}n V. and Simen, Anton and Solano, Enrique and Hegade, Narendra N.},
  journal={arXiv preprint arXiv:2505.08663},
  year={2025}
}

@article{tuziemski2025runtime,
  title={Recent Quantum Runtime (Dis)Advantages},
  author={Tuziemski, Jaros{\l}aw and Paw{\l}owski, Krzysztof and Tarasiuk, Tomasz and Pawela, {\L}ukasz and Gardas, Bart{\l}omiej},
  journal={arXiv preprint arXiv:2510.06337},
  year={2025}
}

@misc{perplexity2025sonar,
  title={Perplexity Sonar API: Model Cards},
  author={{Perplexity AI}},
  year={2025},
  howpublished={\url{https://docs.perplexity.ai/guides/model-cards}}
}

@misc{google2025gemini,
  title={Gemini API Documentation},
  author={{Google}},
  year={2025},
  howpublished={\url{https://ai.google.dev/gemini-api/docs}}
}

@misc{litellm2024,
  title={LiteLLM: A Unified Interface for LLM APIs},
  author={{BerriAI}},
  year={2024},
  howpublished={\url{https://github.com/BerriAI/litellm}}
}

@misc{kipuquantum2025team,
  author       = {{Kipu Quantum GmbH}},
  title        = {Our Team},
  year         = {2025},
  howpublished = {\url{https://kipu-quantum.com/about/our-team/}},
  note         = {Accessed: 2026-03-24}
}

@misc{ibm2025iskay,
  author       = {{IBM}},
  title        = {Kipu Optimization},
  year         = {2025},
  howpublished = {\url{https://quantum.cloud.ibm.com/docs/en/guides/kipu-optimization}},
  note         = {Accessed: 2026-03-24}
}

@misc{kipuquantum2023seed,
  author       = {{Kipu Quantum GmbH}},
  title        = {10.5 Million {EUR} for {G}erman Quantum Software Company {K}ipu {Q}uantum},
  year         = {2023},
  howpublished = {\url{https://kipu-quantum.com/knowledge-hub/press-releases/105-million-eur-for-german-quantum-software-company-kipu-quantum/}},
  note         = {Accessed: 2026-03-24}
}

@misc{kipuquantum2024anaqor,
  author       = {{Kipu Quantum GmbH}},
  title        = {Kipu Quantum Acquires Quantum Computing Platform Built by {A}naqor {AG}},
  year         = {2024},
  howpublished = {\url{https://kipu-quantum.com/knowledge-hub/press-releases/kipu-quantum-acquires-quantum-computing-platform-built-by-anaqor-ag-to-accelerate-development-of-industrially-relevant-quantum-solutions/}},
  note         = {Accessed: 2026-03-24}
}

@article{chandarana2025hsqc,
  title={Hybrid Sequential Quantum Computing},
  author={Chandarana, Pranav and Cadavid, Alejandro Gomez and Romero, Sebasti{\'a}n V. and Simen, Anton and Solano, Enrique and Hegade, Narendra N.},
  journal={arXiv preprint arXiv:2510.05851},
  year={2025}
}

@article{chandarana2026quest,
  title={The Quest for Quantum Advantage in Combinatorial Optimization: End-to-end Benchmarking of Quantum Solvers vs. Multi-core Classical Solvers},
  author={Chandarana, Pranav and Cadavid, Alejandro Gomez and Solano, Enrique and Koch, Thorsten and Woerner, Stefan and Hegade, Narendra N.},
  journal={arXiv preprint arXiv:2603.13607},
  year={2026}
}

@article{farre2025comparing,
  title={Comparing Quantum Annealing and {BF-DCQO}},
  author={Farr{\'e}, Pau and Ordog, Erika and Chern, Kevin and McGeoch, Catherine C.},
  journal={arXiv preprint arXiv:2509.14358},
  year={2025}
}

@misc{kipuquantum2022seed,
  author       = {{The Quantum Insider}},
  title        = {Kipu Quantum Emerges From Stealth, Closes a \texteuro3 Million Funding Round},
  year         = {2022},
  howpublished = {\url{https://thequantuminsider.com/2022/09/15/kipu-quantum-emerges-from-stealth-closes-a-e3-million-funding-round/}},
  note         = {Accessed: 2026-03-24}
}

@misc{anthropic2025claudecode,
  author       = {{Anthropic}},
  title        = {Claude Code: An Agentic Coding Tool},
  year         = {2025},
  howpublished = {\url{https://docs.anthropic.com/en/docs/claude-code}},
  note         = {Accessed: 2026-03-25}
}

@misc{steinberger2025openclaw,
  author       = {Steinberger, Peter},
  title        = {{OpenClaw}: Your Open-Source Personal {AI} Assistant},
  year         = {2025},
  howpublished = {\url{https://openclaw.ai}},
  note         = {Accessed: 2026-03-25}
}

@article{xu2026agentskills,
  author  = {Xu, Renjun and Yan, Yang},
  title   = {Agent Skills for Large Language Models: Architecture, Acquisition, Security, and the Path Forward},
  journal = {arXiv preprint arXiv:2602.12430},
  year    = {2026}
}

@misc{anthropic2025claudecodeskills,
  author       = {{Anthropic}},
  title        = {Claude Code Skills},
  year         = {2025},
  howpublished = {\url{https://docs.anthropic.com/en/docs/claude-code/skills}},
  note         = {Accessed: 2026-03-25}
}

@techreport{mankins1995trl,
  author      = {Mankins, John C.},
  title       = {Technology Readiness Levels: A White Paper},
  institution = {NASA, Office of Space Access and Technology},
  year        = {1995}
}

\end{document}